\documentclass[final]{cvpr}

\usepackage{times}
\usepackage{epsfig}
\usepackage{graphicx}
\usepackage{amsmath}
\usepackage{amssymb}
\usepackage{diagbox}
\usepackage{color,soul,colortbl}
\usepackage{url} 
\usepackage{tabularx}
\usepackage{multirow}
\usepackage{booktabs}
\usepackage{xcolor}
\usepackage{capt-of}
\usepackage[lined,boxed,commentsnumbered,ruled,linesnumbered]{algorithm2e}
\usepackage{pifont}

\newcommand{\cmark}{\ding{51}}%
\newcommand{\xmark}{\ding{55}}%
\graphicspath{{figs/}}
\usepackage[pagebackref=true,breaklinks=true,colorlinks,bookmarks=false]{hyperref}

\definecolor{lblue}{RGB}{13, 152, 255}
\definecolor{lred}{RGB}{255, 108, 108}
\definecolor{dblue}{RGB}{0, 112, 192}
\definecolor{dred}{RGB}{192, 0, 0}
\definecolor{Gray}{gray}{0.9}
\newcommand{\RN}[1]{%
	\textup{\lowercase\expandafter{\it \romannumeral#1}}%
}
\def\confYear{CVPR 2021}
\begin{document}

\title{ Unpaired Image-to-Image Translation via Latent Energy Transport}

\author{Yang Zhao \quad Changyou Chen \\
University at Buffalo, SUNY \\
{\tt\small \{yzhao63, changyou\}@buffalo.edu}
}

\maketitle

\begin{abstract}
Image-to-image translation aims to preserve source contents while translating to discriminative target styles between two visual domains. Most works apply adversarial learning in the ambient image space, which could be computationally expensive and challenging to train. In this paper, we propose to deploy an energy-based model (EBM) in the latent space of a pretrained autoencoder for this task. The pretrained autoencoder serves as both a latent code extractor and an image reconstruction worker. Our model, \textbf{LETIT}\footnote{\textbf{L}atent {\bf E}nergy {\bf T}ransport for {\bf I}mage {\bf T}ranslation}, is based on the assumption that two domains share the same latent space, where latent representation is implicitly decomposed as a content code and a domain-specific style code. Instead of explicitly extracting the two codes and applying adaptive instance normalization to combine them, our latent EBM can implicitly learn to transport the source style code to the target style code while preserving the content code, an advantage over existing image translation methods. This simplified solution is also more efficient in the one-sided unpaired image translation setting. Qualitative and quantitative comparisons demonstrate superior translation quality and faithfulness for content preservation.
Our model is the first to be applicable to 1024$\times$1024-resolution unpaired image translation to the best of our knowledge. Code is available at \url{https://github.com/YangNaruto/latent-energy-transport}.
\end{abstract}

\begin{figure*}[ht!]
    \centering
    \includegraphics[width=\textwidth]{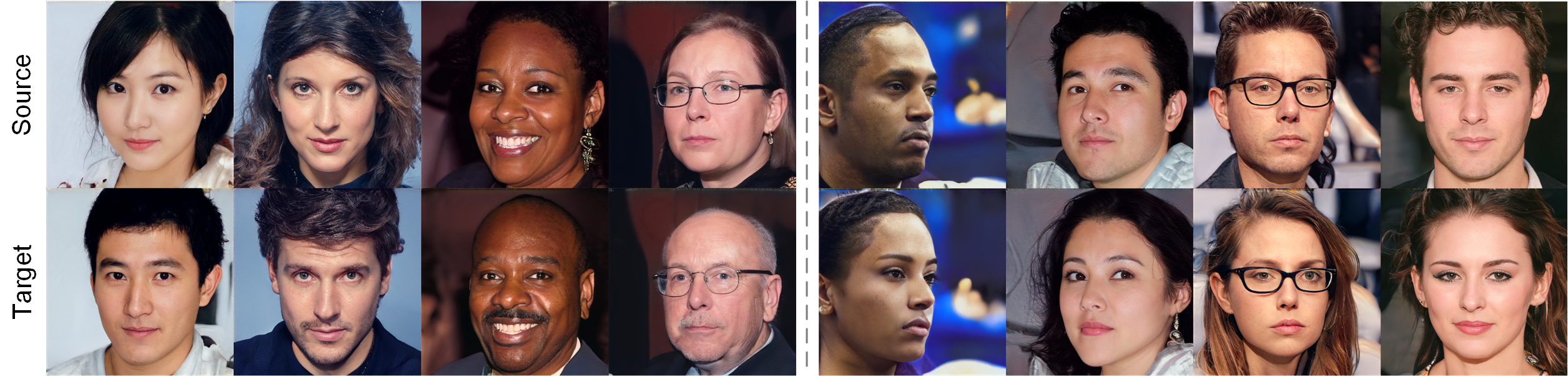}
    \caption{Unpaired image-to-image translation on 1024$\times$1024-resolution images. (\textit{Left}) female to male; (\textit{Right}) male to female.}
    \label{fig:intro-figure}
\end{figure*}

\section{Introduction}
The unpaired image-to-image translation aims to learn pairwise domain mappings without being aware of any paired-image information. Suppose a task of translating between two domains of male and female, denoted as $\mathcal{X}$ and $\mathcal{Y}$ and illustrated in Figure~\ref{fig:intro-figure}. Ideally, one should be able to retain the {\it shared contents}, {\it e.g.}, the irrelevant background and the rough facial skeleton, and only focus on transferring {\it discriminative styles}, {\it e.g.}, hair and beard. Most existing models adopt generative adversarial nets (GANs) \cite{goodfellow2014generative,isola2017image} to enforce the translated style of source instances to be indistinguishable from that of the target domain, which typically relies on an explicit cycle consistency regularizer \cite{yi2017dualgan,zhu2017unpaired,kim2017learning} to maintain the content. However, the enforced cycle consistency is often restrictive, and the learning of two roughly invertible mappings, to some extent, can hinder model optimization efficiency. CUT \cite{park2020contrastive} resorts to contrastive learning as an alternative in the one-sided translation setting. Still, GAN-based approaches need to learn at least one set of an encoder-decoder structured generator and an encoder-based discriminator, which is usually computationally expensive to train \cite{zhao2021learning}.

Apart from GAN-based solutions, CF-EBM~\cite{zhao2021learning}, one of the most recent works, applies the energy-based model (EBM) to realize implicit image translation by direct maximum likelihood estimation (MLE). However, EBM learning leverages the Langevin dynamics for Markov Chian Monte Carlo (MCMC) sampling in the ambient data space, which usually distorts image pixels and is challenging to scale up. Besides, it is unclear whether EBM can learn a disentangled representation of the content and style for better image translation \cite{lee2018diverse}.

To overcome the above issues, we propose a plug-and-play EBM-based model in the latent space. Specifically, we first pretrain an autoencoder (or use an existing one) and then plug the EBM into the latent space to manipulate the extracted latent code to realize image translation. Our latent EBM models an explicit density distribution of latent variables by training a bottom-up latent energy function, which always assigns lower energy to latent variables of the target domain data and higher energy to those of the source domain data. Hence, when sampling from the EBM via Langevin dynamics, the energy gradient (score function) describes a path to transport the latent codes from source to target domains. Most interestingly, we demonstrate that the score function can implicitly and automatically separate both the content and style codes from the whole latent embedding. Thus image translation corresponds to simply evolving the style code. As can be seen in Figure~\ref{fig:intro-figure}, only the style appearance is translated while the content information (the background) is preserved.

Extensive experiments and analysis show that our contributions can be summarized in the following five aspects:
\begin{itemize}
    \item \textbf{Universality:} Our plug-and-play of latent EBM is a universal framework, which can be applied to most autoencoders without introducing any auxiliary networks and engineered loss functions.
    \item \textbf{Efficiency:} Our latent EBM learning is extraordinarily efficient. For example, after pretraining an autoencoder, a very light-design choice of multiple layer perceptron for the latent EBM is adequate to outperform state-of-the-art methods, with only less than one thousand iterations for learning. 
    \item \textbf{Transferability:} A pretrained autoencoder can be reused for multiple image translation datasets, even when pretrained on the human facial dataset and aiming to translate between apples and oranges.
    \item \textbf{Scalability:} Without being restricted by the U-shape design of GANs, we can effortlessly scale the model up for translating images of $1024\times1024$ resolution, as demonstrated in Figure~\ref{fig:intro-figure}.
    \item \textbf{Faithfulness:} Our model can learn faithful translation mappings, where a translated image not only is stylized but also preserves the original content. Take the third column of Figure~\ref{fig:intro-figure} for instance, the facial color and background are perfectly preserved when translating from female to male.
\end{itemize}

The rest of paper is organized as follows. Section~\ref{sec:related-work} reviews and differentiates related works. Section~\ref{sec:ebm} outlines the EBM learning preliminaries. Section~\ref{sec:model} describes our proposed method in detail. Section~\ref{sec:experiments} presents extensive experiments on various kinds of autoencoders to validate our approach. Section~\ref{sec:conclusion} concludes our work and highlights some future research directions.

\section{Related Work}\label{sec:related-work}
\subsection{Image-to-Image Translation}
The solution to unpaired image-to-image translation is usually decomposed into two cooperative paths, the domain-level distribution style matching and the instance-level content preservation. Current notable research efforts can be broadly categorized into two types: GAN-based model and energy-based model.

\paragraph{GAN-based Models}  
MUNIT~\cite{huang2018multimodal}, DRIT~\cite{lee2018diverse}, U-GAT-IT~\cite{Kim2020U-GAT-IT} and StarGAN~\cite{choi2018stargan,choi2020stargan}. 
However, most of these works need to leverage cycle consistency to constrain the domain mapping and enforce the content to be unchanged.  The cycle consistency regularizes the training by reconstructing an original image from its backward translated image at the instance level. Thus, these schemes generally require a combination of generators, discriminators or style encoders~\cite{lee2018diverse,choi2018stargan,choi2020stargan,zhao2020feature}, and complicate engineered loss designs. 
Efforts for one-sided unpaired image-to-image translation also have been made, {\it e.g.}, DistanceGAN~\cite{benaim2017one}, GcGAN~\cite{fu2019geometry} and CUT~\cite{park2020contrastive}. These models apply geometry constraints or contrastive constraints to improve cycle consistency. 

\paragraph{Energy-based Model}
Recently, energy-based generative models have drawn significant attention \cite{xie2016theory, xie2018cooperative,du2019implicit,nijkamp2019learning,deng2020residual,gao2020flow,yu2020training,xie2021learning}. A recent work CF-EBM~\cite{zhao2021learning} demonstrates EBM as a powerful tool to simplify the conditional image generation problem for image translation. In CF-EBM, the source data is taken as the condition, which moves along the energy decay direction via Langevin dynamics~\cite{Langevin1908,welling2011bayesian}. The aforementioned two-level matching is implicitly integrated through MLE-based EBM learning. 

Our proposed approach differs from all the above methods. We summarize the learning schemes in Table~\ref{tab:model-cmp}. Though our practice falls into the EBM learning scheme, it is defined in the high-level lower-dimensional latent space rather than the low-level high-dimensional data space. In this way, the learned latent EBM is better at capturing domain discrepancy to facilitate conditional learning with much fewer iterations because the MCMC sampling is much more efficient. 
\begin{table}[!htbp]
    \begin{center}
    \scalebox{0.8}{
    \begin{tabular}{c|c|c|c}
    \toprule
        Approaches &  Distribution & Instance & Transferability\\ \midrule
        CycleGAN~\cite{zhu2017unpaired} & adversarial & cycle & \xmark \\
        DRIT~\cite{lee2018diverse} & adversarial & cycle & \xmark \\ 
        (M)UNIT~\cite{liu2017unsupervised,huang2018multimodal} & adversarial & cycle &  \xmark \\
        StarGAN~\cite{choi2018stargan,choi2020stargan} & adversarial & cycle & \xmark \\
        Distance~\cite{benaim2017one} & adversarial & cycle+distance & \xmark \\
        CUT~\cite{park2020contrastive} &  adversarial & contrastive & \xmark \\
        CF-EBM~\cite{zhao2021learning} & mle & implicit & \xmark\\
        \rowcolor{Gray}
        Ours & mle (latent) & implicit &\cmark\\
        \bottomrule
    \end{tabular}
    }
    \end{center}
    \caption{Feature-by-feature comparisons of unpaired image-to-image translation models. }
    \label{tab:model-cmp}
\end{table}
\subsection{Autoencoders}
To extract the latent code, we adopt the autoencoder, where an encoder is used to encode an image into a latent code, and a decoder then reconstructs it back to the image space. The recent focus, on the one hand, has been on providing a probabilistic manner to predict the posterior distribution over the latent variables such that the autoencoder is turned into a competitive generative model, {\it e.g.} Variational AE (VAE)~\cite{kingma2014auto}, Vector Quantized VAE (VQ-VAE)~\cite{van2017vqvae,razavi2019generating} and NVAE~\cite{vahdat2020nvae}. On the other hand, the unsupervised disentanglement representation learning is trending, {\it e.g.} $\beta$-VAE~\cite{higgins2016beta,Higgins2017betaVAELB}, Factor-VAE~\cite{kim2018disentangling}, Guided-VAE~\cite{ding2020guided}, TCVAE~\cite{kim2018disentangling} and Adversarial Latent AE~\cite{pidhorskyi2020adversarial}. The goal is to learn factorized and interpretable latent representations that can encode different generative factors, {\it e.g.}, hair, gender, and age in the human faces dataset. 
We refer readers to an excellent repository for more details on this topic\footnote{\url{https://github.com/matthewvowels1/Awesome-VAEs}}. 
This paper validated our approach on different AE variations, including a vanilla VQ-VAE-2~\cite{razavi2019generating} without any generative capability and disentangled generative AEs with other objectives, {\it e.g.}, $\beta$-VAE and ALAE.

\section{MCMC-based Maximum Likelihood Learning of EBM}\label{sec:ebm}
Given an observed image $x \in \mathbb{R}^D$ sampled from a data distribution $p_{data}$, an energy-based model follows:
\begin{align}
    p_\theta(x) = \frac{1}{Z(\theta)}\exp (-E_\theta(x)),
\end{align}
\noindent where $E_\theta(x)$: $\mathbb{R}^D \rightarrow \mathbb{R}$ is the scalar energy function parameterized by $\theta$ and $Z(\theta)=\int \exp (-E_\theta(x))q(x)dx$ is the intractable partition function. Given $N$ observed data points $\{ x_i\}_{i=1}^N$ from the data distribution, the model can be trained by maximizing the log-likelihood  $L(\theta) = \frac{1}{N}\sum_{i=1}^N \log p_\theta(x_i) \approx \mathbb{E}_{x \sim p_{data}} \log (p_\theta (x)).$ The derivative of the negative log-likelihood is
\begin{align}\label{eq:obj}
    -\frac{\partial L(\theta)}{\partial \theta} = \mathbb{E}_{x \sim p_{data}}[\frac{\partial}{\partial \theta}E_\theta(x)] - \mathbb{E}_{\tilde{x} \sim p_\theta}[\frac{\partial}{\partial \theta}E_\theta(\tilde{x})],
\end{align}
\noindent where the second expectation term under $p_{\theta}$ is intractable. We will approximate it via MCMC such that the EBM can be updated by gradient descent.
To sample $\tilde{x} \sim p_\theta$ via MCMC, we rely on Langevin dynamics that recursively computes the following step
\begin{align}\label{eq:langevin}
    \tilde{x}^{t+1} = \tilde{x}^t - \frac{\eta_t}{2}\frac{\partial}{\partial \tilde{x}^t} E_\theta(\tilde{x}^t)+ \sqrt{\eta^t} \epsilon^t, \; \epsilon^t \sim \mathcal{N}(0, \mathbf{I}),
\end{align}
\noindent where $\eta^t$ is the step size typically with polynomially decay to ensure convergence~\cite{welling2011bayesian}; $\epsilon^t$ is a Gaussian sample to capture the data uncertainty and ensure sample convergence.

\section{The Proposed Framework}\label{sec:model}
\subsection{Model}
From a probabilistic perspective, given two image domains $\mathcal{X}$ and $\mathcal{Y}$, our goal of unpaired image translation is to infer a joint distribution based on the marginal distributions $P_\mathcal{X}(x)$ and $P_\mathcal{Y}(y)$ without awareness of paired instances of the two domains. We impose some desired properties on the joint distribution space.

\paragraph{Assumption:} 
Suppose we want to translate from $\mathcal{X}$ to $\mathcal{Y}$. We can achieve this by performing image-to-image translation in both the ambient data space and latent space:

\noindent$(\RN{1})$ In the data space, the goal can be achieved in a straightforward way to learn an image-to-image mapping $F$: $\mathcal{X} \rightarrow \mathcal{Y}$. The mapping $F$ should satisfy two conditions: content preservation and style transfer.

\noindent$(\RN{2})$ Let the associated two latent spaces corresponding to $\mathcal{X}$ and $\mathcal{Y}$ be $\mathcal{Z}_\mathcal{X}$ and $\mathcal{Z}_\mathcal{Y}$ respectively. For our purpose, a latent code should contain both the latent content and latent style information. We thus formulate the problem as learning a mapping $G$: $\mathcal{Z}_\mathcal{X} \rightarrow \mathcal{Z}_\mathcal{Y}$ such that it satisfies two conditions: latent content preservation and latent style transfer.

As we claimed before, most existing works are based on the first framework, which is computationally expensive in training and requires complicated loss design to satisfy both conditions. Our approach is based on the second framework. 
As the latent space is an abstract-level and compact representation of the data, it is reasonable to assume that two domains $\mathcal{X}$ and $\mathcal{Y}$ share the same latent space $\mathcal{Z}$, with each code $z$ decomposed into a content code $c$ and a domain-specific code $s_\mathcal{X}$ for $\mathcal{X}$ or $s_\mathcal{Y}$ for $\mathcal{Y}$, {\it i.e.}, $z = [c, s_{\mathcal{X}}]$ or $z = [c, s_{\mathcal{Y}}]$. For a source instance $x$ with a latent code $z_x=[c, s_x]$, the translation aims at only transforming the source style code $s_x$ to the target style code space $s_\mathcal{Y}$. We will demonstrate emperically that our model achieve automatically learn to do this without explicitly specifying out the content code and domain-specific code.

Note our setting is essentially different from the purpose of the partially shared latent space assumption in MUNIT~\cite{huang2018multimodal}, which aims to ease the learning of a pair of underlying encoder and decoder. In our model, our assumption enables one to directly learn the latent code mapping between two domains to realize image translation. 


\paragraph{Method:} 
As described above, we aim to transport latent codes via EBM directly and postulate (but verify empirically) that the latent EBM can transport the source style code $s_x$ into the target style space $s_\mathcal{Y}$ while preserving source content codes $c$. To achieve this, one needs an encoder to extract the latent code and a decoder to reconstruct the target image. 
\begin{figure}[!htbp]
    \centering
    \includegraphics[width=0.9\linewidth]{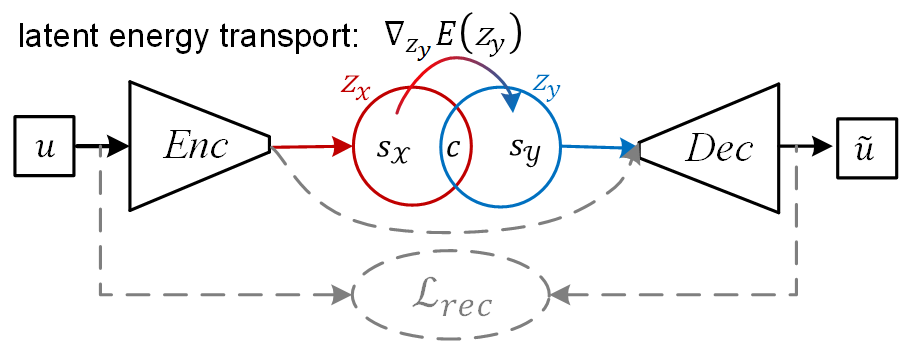}
    \caption{The proposed model: data flow and energy transport. The dashed data flow only exists in the autoencoder pretraining stage; $u$ could be either $x$ or $y$.}
    \label{fig:model}
\end{figure}
The model is illustrated in Figure~\ref{fig:model}. Specifically, we consider a pretrained autoencoder, including an encoder $Enc(\cdot)$ and a decoder $Dec(\cdot)$. The pretraining procedure follows:
\begin{align}
    &\text{Encoding}: z=Enc(u), u \sim P_\mathcal{X} \cup P_\mathcal{Y} \\\nonumber
    &\text{Decoding}: \tilde{u}=Dec(z) \\ 
    &\text{Reconstruction loss}: \mathcal{L}_{rec} = \mathbb{E}_{u \sim P_\mathcal{X} \cup P_\mathcal{Y}} ||u-\tilde{u}||_2 \nonumber
\end{align}
 Suppose the task is to adapt $P_\mathcal{X}$ to $P_\mathcal{Y}$, we aim to learn an EBM $E_{x\rightarrow y}$ satisfying:
\begin{align}
    p_\theta(z_y) = \frac{1}{Z(\theta)}\exp (-E_{x\rightarrow y}(z_y)), z_y=Enc(y).
\end{align}
The learning process of $E_{x\rightarrow y}$ is very simple by adopting \eqref{eq:langevin}. To sample from the EBM $E_{x\rightarrow y}$, we modify \eqref{eq:langevin} as:
\begin{align}\label{eq:langevin-i2i}
    \tilde{z}_{y}^{t+1} = \tilde{z}_{y}^t - \frac{\eta^t}{2}\frac{\partial}{\partial \tilde{z}_{y}^t} E_{x\rightarrow y}(\tilde{z}_{y}^t)+ \sqrt{\eta^t} \epsilon^t,
\end{align}
where $\tilde{z}_{y}^0 = z_x = Enc(x) \; \text{and} \; x \sim P_\mathcal{X}$. The optimization of $E_{x\rightarrow y}$ exactly follows \eqref{eq:obj}. After $T$ Langevin steps, the reconstructed $Dec(z_{y}^T)$ will serve as the translation of $x$. Note the above learning only requires optimization of the EBM; thus, it is very computationally efficient. 

In practice, it might happen that reconstructions from the decoder exhibit a blurry effect. Existing works introduce an additional EBM to refine the reconstructed output. This approach has proved successful in the recent proposed VAEBM~\cite{xiao2020vaebm}. \cite{pang2020learning} also provides a promising method to learn a generator and the latent EBM cooperatively. By contrast, our model does not consider this extra effort, as we want to focus on the simple plug-and-play setting, and our results already demonstrate excellent image quality.

\subsection{Proof-of-Concept Verification}
\begin{figure}[!htbp]
    \begin{center}
        \includegraphics[width=\linewidth]{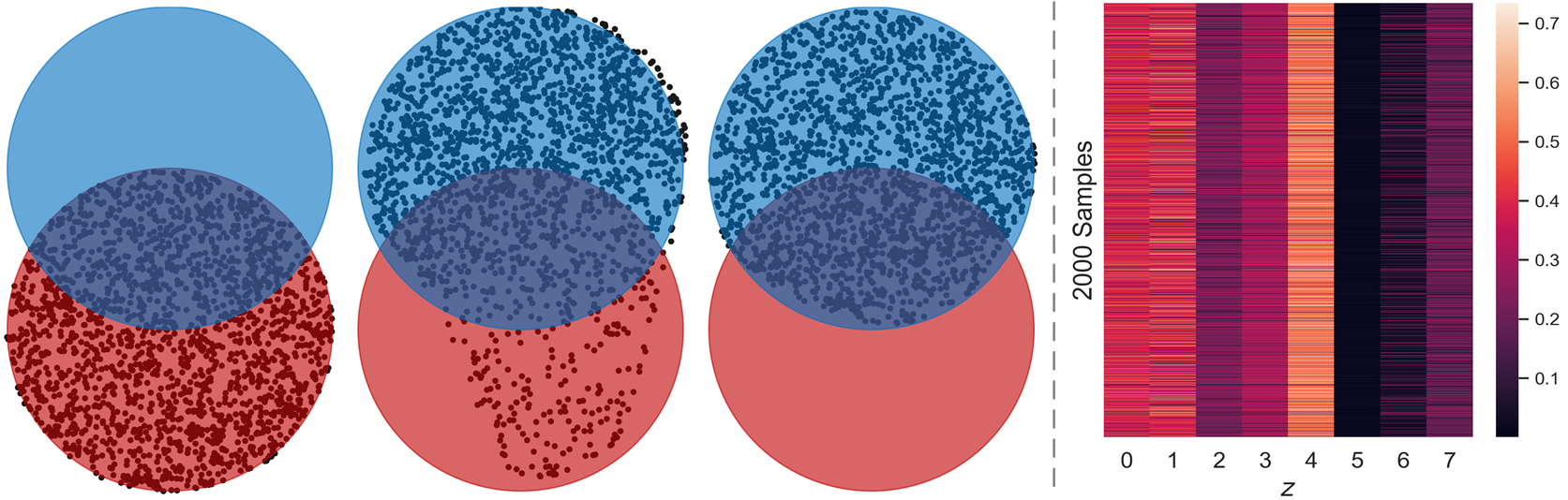}
    \end{center}
    \caption{Domain evolving process from \textit{\textcolor{lred}{red pie}} to \textit{\textcolor{lblue}{blue pie}} on synthetic data. (\emph{Left}) Latent code transition corresponding to epochs 0, 5 and 50. (\emph{Right}) The latent code shift after 10 Langevin steps.}
    \label{fig:latent-syn}
\end{figure}

We conduct a proof-of-concept verification on our assumption with two synthetic domains, the \textit{\textcolor{lred}{red pie}} and the \textit{\textcolor{lblue}{blue pie}} as shown in Figure~\ref{fig:latent-syn}. Following the proposed approach, we firstly pretrain until convergence a one-layer autoencoder and set the latent dimension to 8. A two-layer EBM is then learned to transport the latent code from one domain to the other using a 10-step MCMC. In Figure~\ref{fig:latent-syn}, we visualize the process of sample transportation from the black dots of the \textit{\textcolor{lred}{red pie}} to the \textit{\textcolor{lblue}{blue pie}} in the data space. We observe that the proposed approach indeed can translate between the domains. To illustrate that our model can also automatically learn to distinguish the content and style codes, we calculate the \textbf{aggregated absolute latent code shift}, which is defined as the sum of absolute gradients during MCMC in \eqref{eq:langevin-i2i}, {\it i.e.}, $\sum_t\|\nabla_{z_y^t}E(z_y^t)\|$. As shown in the heatmap of Figure~\ref{fig:latent-syn}, the model does not learn a uniform shift among all latent dimensions. Instead, the 5-$th$ latent dimension of most samples exhibits almost zero shift, meaning that this dimension can be considered as the shared content code. 

\section{Experiments}\label{sec:experiments}
We evaluate our approach across several image-to-image translation datasets with three autoencoder structures.
\paragraph{Autoencoders:} We adopt three autoencoder structures, with key characterizations listed in Table~\ref{tab:base-cmp}.
\begin{itemize}
    \item \textbf{$\beta$-VAE}~\cite{Higgins2017betaVAELB}: It is a well-known unsupervised disentangle representation learning model. It modifies the objective -- the evidence lower bound (ELBO), of variational autodencoder~\cite{kingma2014auto} and enforces the latent bottleneck (weighted Kullback–Leibler (KL) divergence) to encourage a \emph{factorized representation}. 
    \item \textbf{ALAE}~\cite{pidhorskyi2020adversarial}: It leverages the adversarial data distribution learning and latent distribution learning in AE. It learns a \emph{descent disentangle representation} and demonstrates \emph{comparable sample quality} with GAN. Thus, it is reasonable to compare it with state-of-the-art image translation models directly. And it is also beneficial for the plug-and-play of much higher-resolution models.
    \item \textbf{VQ-VAE-2}~\cite{razavi2019generating}: We treat it as a vanilla autoencoder without generative ability if it is trained only with the reconstruction loss and the second-stage PixelCNN prior training is discarded \cite{van2016conditional}. Hence, the decoder can only remember specific latent codes but not learn a distribution mapping like that of a generator. The reasons why we consider it is: ($\RN{1}$) the latent space is built on several feature maps which are more challenging than 1-d vector; ($\RN{2}$) the reconstruction performance is in the lead; ($\RN{3}$) most importantly, we use VQ-VAE-2 to demonstrate the strong \emph{transferability} of image translation via latent energy transport after it is pretrained on a large-scale or essentially irrelevant dataset.
\end{itemize}
\begin{table}[!htbp]
    \begin{center}
    \scalebox{0.75}{
    \begin{tabular}{c|c|c|c|c}
    \toprule
        Models & Data & Latent & Generative? & Disentangle? \\ \midrule
        $\beta$-VAE &  ELBO* & KL & \cmark & \cmark \\
        ALAE & Adversarial & MSE & \cmark& \cmark \\
        VQ-VAE-2 &  MSE & Quantization & \xmark & \xmark\\
        \bottomrule
    \end{tabular}
    }
    \end{center}
    \caption{Comparison of the three autoencoder models.}
    \label{tab:base-cmp}
\end{table}
\vspace{-4mm}
\paragraph{Datasets: } 
For $\beta$-VAE, we apply the CelebFaces Attributes Dataset (CelebA) ~\cite{liu2015faceattributes} with over 200K celebrity facial images. The dataset is also widely used in disentangled representation learning. We resize the images into $64\times64$ resolution and divide the dataset into {\it male} and {\it female} according to the gender attribute for translation.

To compare our model with StarGAN v2, where we directly adopt the two high-fidelity image-to-image translation datasets that are used in its experiment: CelebA-HQ~\cite{karras2018progressive} and Animal Faces (AFHQ)~\cite{choi2020stargan}. ($\RN{1}$) CelebA-HQ: it contains 30k celebrity facial images, which are manually split into 17,943 female faces and 10,057 male faces for training~\cite{choi2020stargan}. The rest 2000 images are evenly divided as testing data. We conduct experiments on both 256$\times$256 and 1024$\times$1024 resolutions. ($\RN{2}$) AFHQ (256$\times$256): it includes 15k animal faces and is evenly distributed into three challenging domains, cat, dog and wildlife. Each domain uses 500 images for testing and the rest for training.

To investigate the generalization capability through the combination of the latent energy transport and VQ-VAE-2, we conduct experiments across a couple of datasets used in CycleGAN~\cite{zhu2017unpaired}: \textit{apple2orange} and two painting style transfer datasets, \textit{vangogh2photo} and \textit{ukiyoe2photo}. More details are given in the Appendix.

For the large-scale pretraining, we use the following two datasets: 
($\RN{1}$) The Flickr-Faces-HQ (FFHQ) dataset~\cite{karras2019style} which consists
of 70k high-quality images (1024$\times$ 1024) with more variations than CelebA-HQ in terms of accessories, age, ethnicity and image background; 
($\RN{2}$) ImageNet-1k~\cite{deng2009imagenet} dataset containing over 1 million images of 1000 distinct classes. The images are resized to 256$\times$256.

\paragraph{Evaluation Metrics: }
We consider two commonly used metrics for evaluating image synthesis quality and human perceptual study for visual quality evaluation. 

$(\RN{1})$ FID~\cite{heusel2017gans} compares the statistics (mean and variances of Gaussian distributions) between generated samples and real samples. FID is consistent with increasing disturbances and human judgment. Lower scores indicate that a model can create higher quality images. 

$(\RN{2})$ KID~\cite{binkowski2018demystifying} improves FID as an unbiased estimator, making it more reliable when fewer test images are available. We use generated images translated from all test images in the source domain vs. test images in the target domain to compute KID. Lower KID values indicate that images are better translated. 

$(\RN{3})$  We use the Amazon Turker (AMT) for human percpetual study.

\subsection{Analysis on $\beta$-VAE}
$\beta$-VAE is learned by maximizing a lower bound:
\begin{align}
    \mathsf{ELBO} = -\mathcal{L}_{rec} - \beta \mathsf{KL}(q(z|u)||p(u))
\end{align}
where $p(z)$ is the Gaussian prior and $q(z|u)$ is the approximate posterior; $\beta$ is the adjustable hyperparameter and when $\beta=1$, it recovers the original VAE \cite{kingma2014auto}. Generally, choosing a different value of $\beta$ leads to a trade-off between the reconstruction quality and the disentangled latent space. 

\begin{figure}[!htbp]
    \centering
    \includegraphics[width=\linewidth]{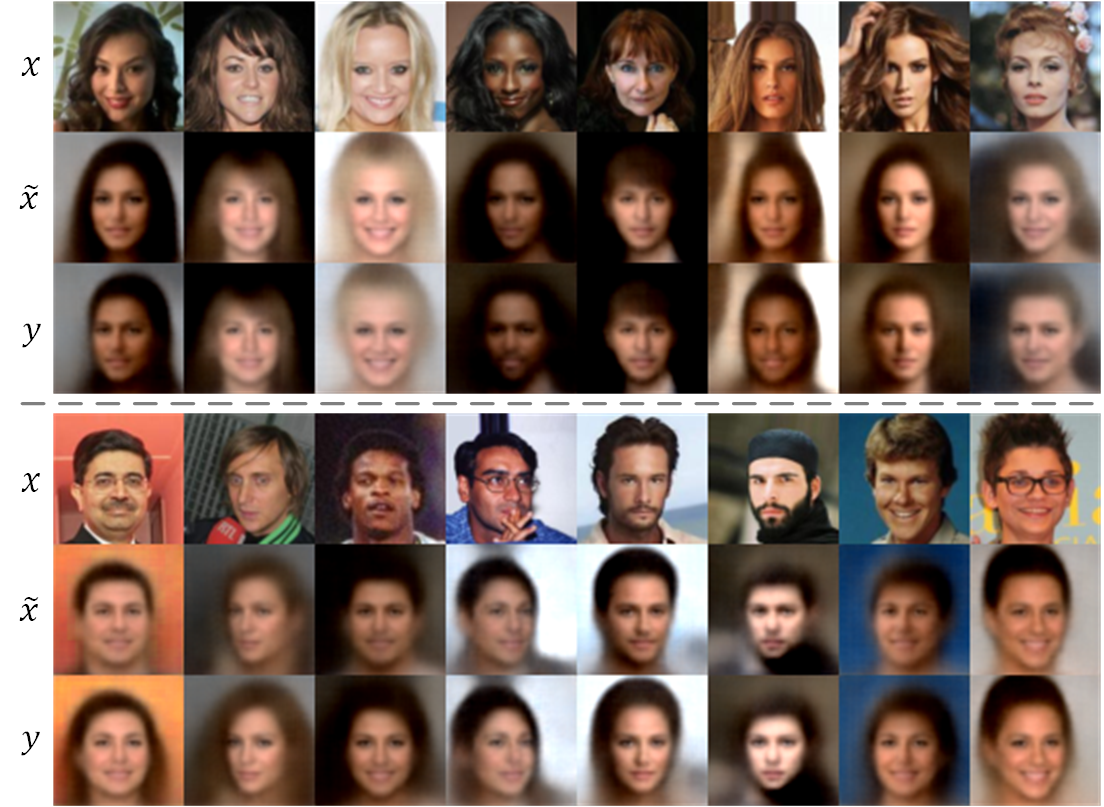}
    \caption{Uncurated image translation results based on $\beta$-VAE in CelebA. $x$: the input, $\tilde{x}$: the reconstruction, $y$: the translated output. ({\it Top}) female to male. ({\it Bottom}) male to female.}
    \label{fig:betavae}
\end{figure}

\paragraph{Implementation}
Since we focus on the latent energy transport aspect for the unpaired image translation task, we simply choose a random $\beta=10$ for evaluation. We set $dim(z)=32$ and construct the EBM with a 1-layer MLP. More detailed experimental settings are given in the Appendix. Note that after the pretraining of $\beta$-VAE\footnote{\url{https://github.com/1Konny/Beta-VAE}}, the latent EBM only requires around 200 additional training iterations (takes 1 minute on TITAN XP) to find a descent translation path on the latent space. Figure~\ref{fig:betavae} shows the results of the two-direction image translation on CelebA. It is observed that the discriminative features, like hair, beard and facial-skeleton, can be effectively uncovered for image translation via latent energy transport. Note we do not expect this model to generate high-quality images as $\beta$-VAE itself is not a good model for this purpose. The other two models will achieve high-quality images. 

\begin{figure}[!htbp]
    \centering
    \includegraphics[width=\linewidth]{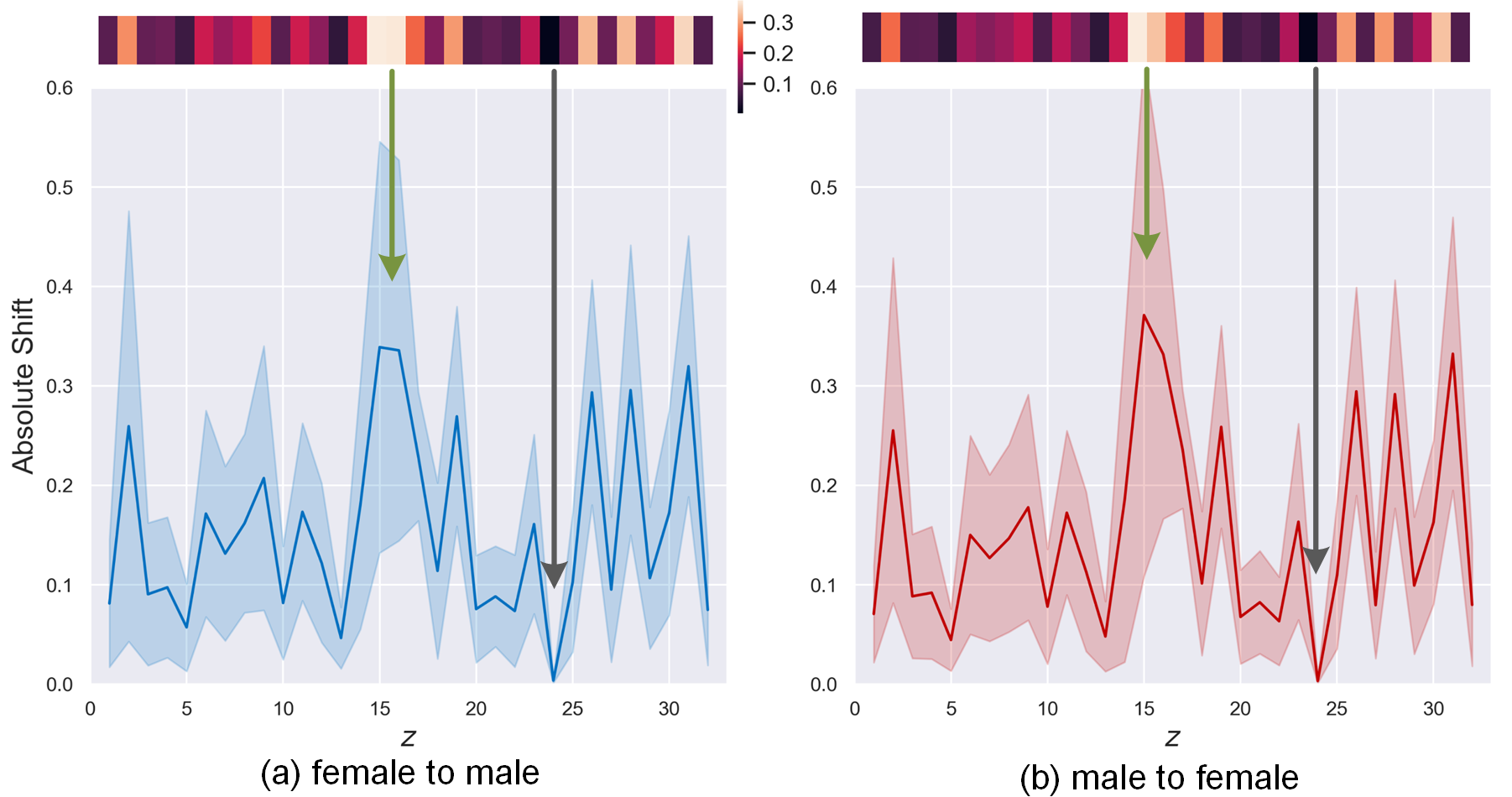}
    \caption{Visualization of the latent code shift via MCMC, which is acquired from 1,000 test images.}
    \label{fig:betavae-latent}
\end{figure}
\paragraph{Behavior of latent space} 
We use this model to visualize the aggregated absolute latent code shift via MCMC, plotted in Figure~\ref{fig:betavae-latent}. We find translations of the two directions reveal nearly identical responses to the same latent dimensions, meaning the models have been likely to learn two almost \emph{mutually inverse} mappings of the two domains, although two EBMs are learned separately without explicitly enforced cycle consistency. Specifically, dimensions 15, 16 (green arrow) are activated, representing style codes, while dimension 24 (gray arrow) stays inactivated, representing content code. Moreover, it is observed that image translation through manipulating the encoded latent code is done by the cooperation of various latent dimensions instead of one single dimension. This experiment validates our assumption and the effectiveness of latent energy transport. 

\begin{figure*}[!htbp]
    \centering
    \includegraphics[width=0.95\textwidth]{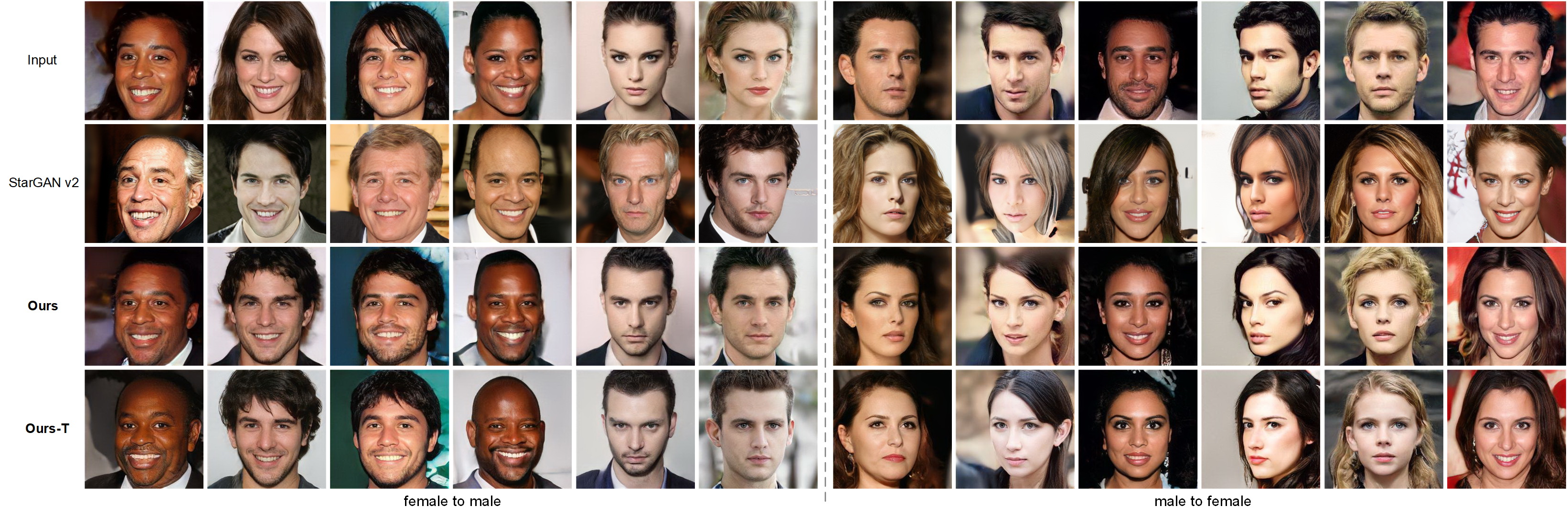}
    \caption{Faithful unpaired image-to-image translation on CelebA-HQ. The third row (\textbf{Ours}) presents the results from the model pretrained on CelebA-HQ whereas the fourth row (\textbf{Ours-T}) gives the transferred translations from the model that is pretrained on FFHQ.}
    \label{fig:celeba-cmp}
\end{figure*}
\begin{figure*}[!htbp]
    \centering
    \includegraphics[width=0.95\textwidth]{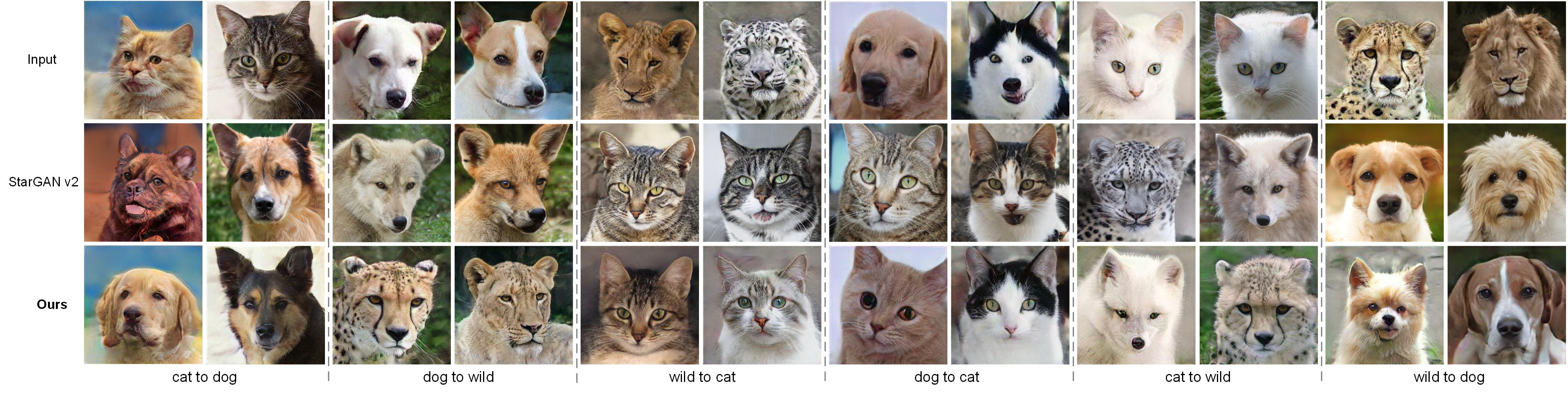}
    \caption{Faithful unpaired image-to-image translation on AFHQ.}
    \label{fig:afhq-cmp}
\end{figure*}

\subsection{Analysis on ALAE}
\paragraph{Implementations} 
We follow all the experiment settings in \cite{pidhorskyi2020adversarial} and pretrain ALAE\footnote{\url{https://github.com/podgorskiy/ALAE}} on CelebA-HQ, AFHQ and FFHQ datasets. We test the unpaired image-to-image translation on CelebA-HQ and AFHQ datasets. The latent dimension of ALAE is 512. The latent EBM has one hidden layer (512-2048-1) activated by LeakyReLU. We find that the performance is not very sensitive to the architecture. To optimize the EBM, we apply the stochastic gradient descent (SGD) with a learning rate 0.1. We set the batch size to 32 and train the EBM for 2000 iterations. We run 20 Langevin steps with a step size of 1.0.

\paragraph{Qualitative results:}
StarGAN v2~\cite{choi2020stargan} is the current state-of-the-art image-to-image translation approach on AFHQ and CelebA-HQ. We show the qualitative comparison between our method and StarGAN v2 in Figure~\ref{fig:celeba-cmp} and Figure~\ref{fig:afhq-cmp}. For StarGAN v2, we feed all the reconstructed images from ALAE for translation and pick the best image from its diverse results. It's fair for us to use the reconstructed image from ALAE as inputs because we are manipulating the latent space with the proposed approach and the change will be reflected in the reconstruction. Figure~\ref{fig:celeba-cmp} presents two variants of outcomes from our approach: one ALAE is pretrained on CelebA-HQ and the other on FFHQ. We observe that both results give faithful and high-fidelity translations. In particular, compared with StarGAN v2, our approach obviously works better at preserving the facial color and backgrounds while translating between males and females. Similar observations are also seen in AFHQ as shown in Figure~\ref{fig:afhq-cmp}.
\begin{table}[!htbp]
    \centering
    \scalebox{0.85}{
    \begin{tabular}{l|c|c}
    \toprule
        Models & CelebA-HQ & AFHQ \\ \midrule
        MUNIT~\cite{huang2018multimodal}  & 31.4 & 41.5 \\
        DRIT~\cite{lee2018diverse}  & 52.1 & 95.6 \\
        MSGAN~\cite{mao2019mode}  & 33.1 & 61.4 \\
        StarGAN v2~\cite{choi2020stargan}  & \textbf{13.7} & \textbf{16.2} \\
        \rowcolor{Gray}
        LETIT (Ours) & \textcolor{blue}{\textbf{12.5}}  & \textcolor{blue}{\textbf{15.9}}\\
         \bottomrule
    \end{tabular}
    }
    \vspace{2mm}
    \caption{Quantitative comparison with baselines. All numbers except for our approach are from StarGAN v2~\cite{choi2020stargan}.}
    \label{tab:comp-stargan}
\end{table}
\begin{table}[!htbp]
    \centering
    \scalebox{0.85}{
    \begin{tabular}{c|c|c|c|c}
    \toprule
    \multirow{2}{*}{Models}  & \multicolumn{2}{c}{CelebA-HQ} & \multicolumn{2}{c}{AFHQ} \\
         &   Quality     &    Faith     &   Quality     &       Faith           \\ \midrule
    StarGAN v2 & 25.6 & 2.2 & 56.3 & 38.4 \\
    \rowcolor{Gray}
    LETIT (Ours) & 35.4 & \textcolor{blue}{\textbf{51.7}} & \textbf{43.7}& \textcolor{blue}{\textbf{61.6}} \\
    \rowcolor{Gray}
    LETIT (Ours-T) & \textcolor{blue}{\textbf{39.0}} & 46.1 & - & - \\ \bottomrule
    \end{tabular}
    }
    \vspace{2mm}
    \caption{Human perceptual study regarding translation quality and faithfulness.}
    \label{tab:amt}
\end{table}
\vspace{-3mm}
\paragraph{Quantitative results:}
We report the FID scores on several models in Table~\ref{tab:comp-stargan}. On both CelebA-HQ and AFHQ datasets, our approach consistently performs better than the baselines. For the AMT perceptual study, each image is judged by six users, who are asked to select the best qualitative translated image with two standards considered: the visual \textbf{Quality} and the \textbf{Faith}ful translation measured by the extent of source content preservation, including background and expression. We inform the participants of the name of the target domain and six example images of the target domain as a visual illustration. The result is shown in Table~\ref{tab:amt}, where our models again outperform StarGAN v2.

\begin{figure*}[!htbp]
    \centering
    \includegraphics[width=\textwidth]{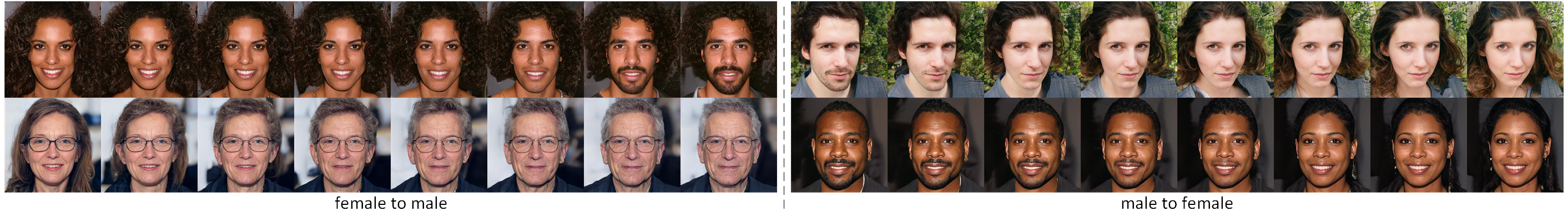}
    \caption{Smooth unpaired image translation dynamics driven by MCMC, on 1024$\times$1024-resolution images.}
    \label{fig:celeba-hq-seqeunce}
\end{figure*}

\vspace{-2mm}
\paragraph{High-resolution translation}
We conduct the image-to-image translation on CelebA-HQ with a resolution of $1024\times 1024$. To the best of our knowledge, this is the first trial on such a high-resolution setting. The autoencoder is pretrained on FFHQ-$1024^2$. Other implementation details are the same as CelebA-HQ, except that the batch size is made to 16 to fit the GPU memory. Since the model only deals with the latent space, the cost is similar as translating $256\times 256$ images. Some visualized results are shown  in Figure~\ref{fig:intro-figure}. We additionally visualize the smooth translation evolution in the data space as the latent code evolves in the latent space. Some example images are shown in Figure~\ref{fig:celeba-hq-seqeunce}, from which we can see how images are smoothly transported to the target domains. More results are included in the Appendix.

\subsection{Analysis on VQ-VAE-2}

\paragraph{Implementations:} 
We use the original VQ-VAE-2 implementation \cite{razavi2019generating} with source code. To accelerate the pretraining and facilitate the latent EBM learning, we make several modifications to the original architecture~\cite{razavi2019generating}, including $(\RN{1})$ We adopt two-level latent maps, bottom and top, such that the EBM is defined on the latent space just before the entrance to the bottom decoder. $(\RN{2})$ We inject layer-wise random Gaussian noise to the decoder for more flexibility. $(\RN{3})$ We adjust the codebook dimension and codebook size. Note that both bottom and top codebooks use the same configuration. The impact on reconstruction quality regarding the Mean Squared Error (MSE) is given in Table~\ref{tab:mse-vqvae}. We set the codebook dimension to 32 and codebook size to 256 such that the latent dimension is $64\times64\times64$~\cite{razavi2019generating}. The detailed latent EBM architecture and training settings are given in Appendix. We compare our model with two one-sided translation models, CF-EBM and CUT.

\begin{figure}[!htbp]
    \centering
    \includegraphics[width=0.85\linewidth]{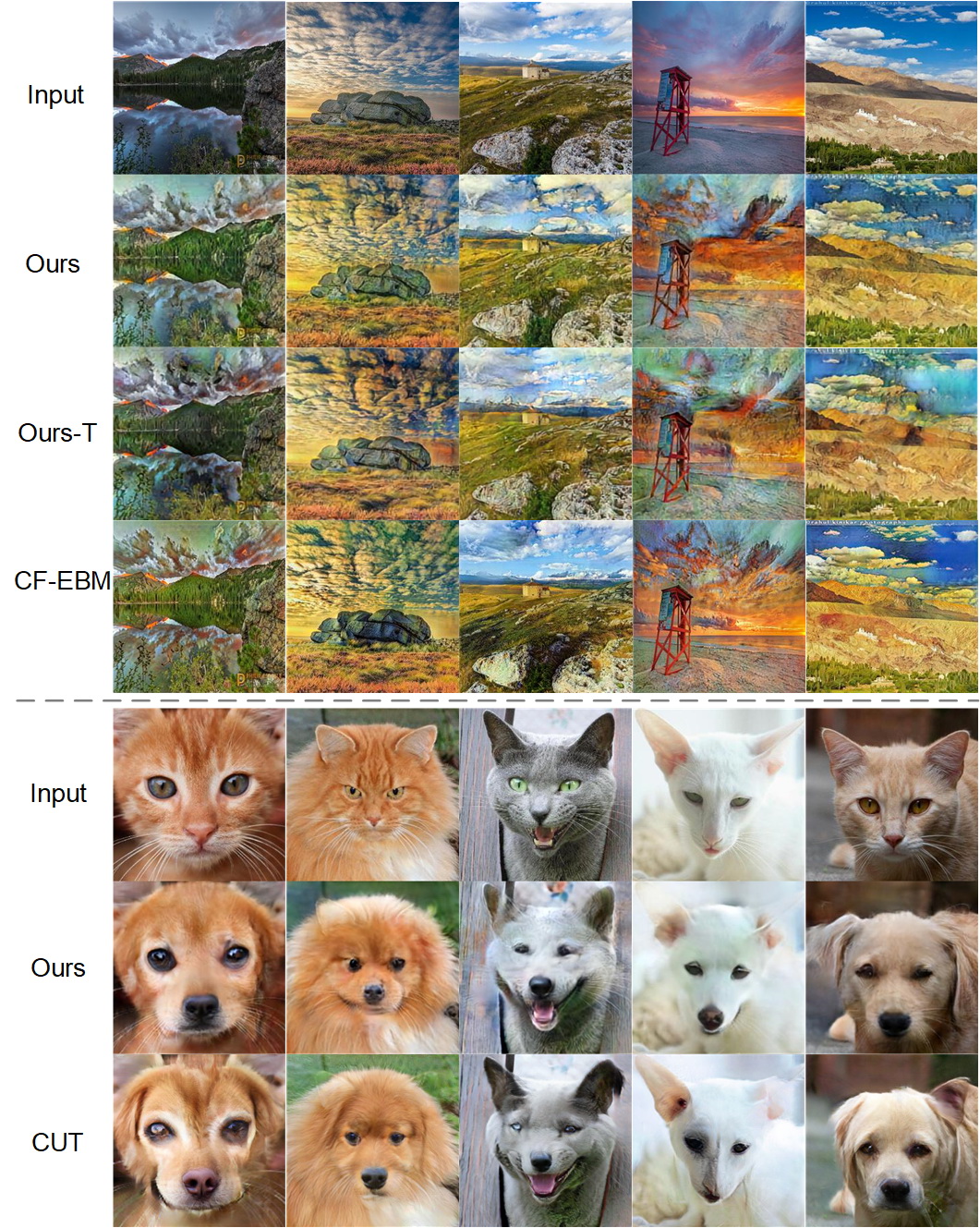}
    \caption{Qualitative comparisons based on VQ-VAE-2. Ours-T means this autoencoder is pretrained on AFHQ.}
    \label{fig:i2i-vqvae}
\end{figure}

\paragraph{Qualitative results:}
Figure~\ref{fig:i2i-vqvae} visualizes some results from different approaches. In the top row of Figure~\ref{fig:i2i-vqvae}, we show the outcomes from different pretrained autoencoders on photo$\rightarrow$vangogh. We observe, although the autoencoder is pretrained on irrelevant datasets, that our model can still generate reasonable translations. However, the most competitive results come from the pretrained autoencoders with the same dataset, which has better color controllability. 
The bottom row in Figure~\ref{fig:i2i-vqvae} compares CUT with our model on AFHQ cat$\rightarrow$dog.
Additional results are given in Appendix.
\vspace{-4mm}
\vspace{-5mm}
\paragraph{Quantitative results:}
We firstly compare our approach with the recent CF-EBM mode~\cite{zhao2021learning} on photo$\rightarrow$vangogh. As shown in Table~\ref{tab:vqvae-cmp}, our approach requires much less time than CF-EBM while achieving comparable KID. We postulate that: $(\RN{1})$ A slightly higher KID is due to the blurry output from the decoder; $(\RN{2})$ The latent code is a high-level compact representation compared with the raw data so that our latent EBM requires less iterations to train. 
CUT is the current stat-of-the-art one-sided image translation approach on AFHQ cat$\rightarrow$dog. We compare the performance including the sample quality and memory requirement. As shown in Table~\ref{tab:vqvae-cmp}, our method achieves greatly better FID score while requiring much less GPU memory. The GPU memory is measured when both methods set the batch size to 1 on GTX 1080Ti.

\begin{table}[!htbp]
    \centering
    \scalebox{0.75}{
    \begin{tabular}{ccccc}\toprule
        \multirow{2}{*}{Models} & \multicolumn{2}{c}{photo $\rightarrow$ vangogh} & \multicolumn{2}{c}{cat $\rightarrow$ dog} \\ \cmidrule(lr){2-3} \cmidrule(lr){4-5}
         & KID$\downarrow$ & Days & FID$\downarrow$ & Mem(GB) \\ \midrule
        CF-EBM~\cite{zhao2021learning} & \textbf{4.25} & 0.6 & 55.1 & 2.50 \\ 
        CUT~\cite{park2020contrastive} & 4.81 & 0.7 & 76.2 & 3.03 \\
        \rowcolor{Gray}
        LETIT (Ours) & 4.42 & \textbf{\textcolor{blue}{0.1}} & \textbf{\textcolor{blue}{45.2}} & \textbf{\textcolor{blue}{1.24}} \\
        \bottomrule
    \end{tabular}
    }
    \vspace{1mm}
    \caption{\small{VQ-VAE-2-based model comparison with CF-EBM and CUT. We only report available numbers from the original papers.}}
    \label{tab:vqvae-cmp}
\end{table}
\vspace{-3mm}
\subsection{Training Cost}
We compare the training cost between our approach and the two strong baselines, StarGAN v2~\cite{choi2020stargan} and CUT~\cite{park2020contrastive}. As seen in Table~\ref{tab:vqvae-cmp}, the time budget for training the latent EBM is incredibly low. Our model is still much more efficient even if the autoencoder training cost counts. Therefore, our approach is a light-weight and practical choice in scenarios where the computational resources are limited.

\begin{table}[!htbp]
    \centering
    \scalebox{0.75}{
    \begin{tabular}{c|c|c|c}
    \toprule
        Models & \#Param(M) & Time(days) & GPU \\ \midrule
        StarGAN v2~\cite{choi2020stargan} & 79 & 3.0 & TITAN V100 \\
        CUT~\cite{park2020contrastive} &  15 & 2.0 & GTX 1080Ti \\ \midrule
        EBM-VQ-VAE-2 (ours) & 17 \scriptsize{\textcolor{cyan}{+1.4}} & 0.6 \scriptsize{\textcolor{cyan}{+0.3}} & GTX 1080Ti \\
        EBM-ALAE (ours) & 54 \scriptsize{\textcolor{cyan}{ +1.6}} & 1.1 \scriptsize{\textcolor{cyan}{ +0.05}} & TITAN XP \\ \bottomrule
    \end{tabular}
    }
    \vspace{1mm}
    \caption{Comparison of computational cost on AFHQ. $a$\textcolor{cyan}{+b} denotes the time for pretraining is $a$, and $b$ for training the EBM.}
    \label{tab:cmp-computation}
\end{table}

\vspace{-3mm}
\section{Conclusion}\label{sec:conclusion}
We propose an efficient and readily plug-and-play approach for one-sided unpaired image-to-image translation, via latent energy transport in a pretrained latent space. The introduced latent EBM learns to implicitly and simultaneously transfer styles and preserve content, without relying on a complicated cycle constraint. Extensive experiments on various autoencoder structures and datasets have demonstrated the strong performance, remarkable efficiency, practical faithfulness and scalability of the proposed universal approach.

\paragraph{Acknowledgments}
This work was supported in part by the Verizon Media FREP program.

\clearpage
\newpage

{\small
\bibliographystyle{ieee_fullname}
\bibliography{egbib}
}

\clearpage
\newpage
\appendix\label{sec:appendix}

\section{Algorithms}
We list the training and translation (sampling) strategy in Algorithm~\ref{algo-train} and Algorithm~\ref{algo-2}, respectively.

\paragraph{Training} To learn the latent energy-based model $E_{x\rightarrow y}$, we take the latent codes $z_y$ of the target domain $P_\mathcal{Y}$ as our ground truth data. The latent codes $z_x$ of the source domain $P_\mathcal{X}$ serve as the initial samples of the latent MCMC as shown in Eq.~\ref{eq:langevin-i2i}. The training algorithm follows:
\begin{algorithm}
\SetAlgoLined
\begin{flushleft}
    \textbf{Input:} source domain $P_\mathcal{X}$, target domain $P_\mathcal{Y}$\\
    \textbf{Output:} latent energy function $E_{x \rightarrow y}$
\end{flushleft}
\vspace{-0.1in}
\While {not converged}{ 
{\small \color{blue} \# Draw source and target domain image $x$ and $y$}\\
$x\sim P_\mathcal{X}, \;y\sim P_\mathcal{Y}$\\
{\small \color{blue} \# Encode sample $\tilde{z}_y^0$ and target $z_y$}\\
$\tilde{z}_y^0 = z_x = Enc(x)$ ,$z_y = Enc(y)$ \\
{\small \color{blue} \# MCMC to sample $\tilde{z}_y^T$}\\
\For {$t=1:T$}{
    Update $\tilde{z}_y^t$ according to Eq.~\ref{eq:langevin-i2i}\\
}
{\small \color{blue} \# Update $E_{x\rightarrow y}$ based on $\tilde{z}_y^T$ and $z_y$}\\
Update  $E_{x\rightarrow y}$ according to Eq.\ref{eq:obj}
}
 \caption{\small{Latent Energy-based Model Training}}
 \label{algo-train}
\end{algorithm}

\paragraph{Translation} Given an input image, the translation process is simple.
\begin{algorithm}
\SetAlgoLined
\begin{flushleft}
    \textbf{Input:} $x$\\
    \textbf{Output:} $y$
\end{flushleft}
\vspace{-0.1in}
$z_y^0 = z_x = Enc(x)$ \\
\For {$t=1:T$}{
    Update $z_y^t$ according to Eq.~\ref{eq:langevin-i2i}\\
}
$y = Dec(z_y^T)$
 \caption{\small{Latent Energy Transport for Translation}}
 \label{algo-2}
\end{algorithm}

\section{$\beta$-VAE}
\begin{figure*}[!htbp]
    \centering
    \includegraphics[width=\textwidth]{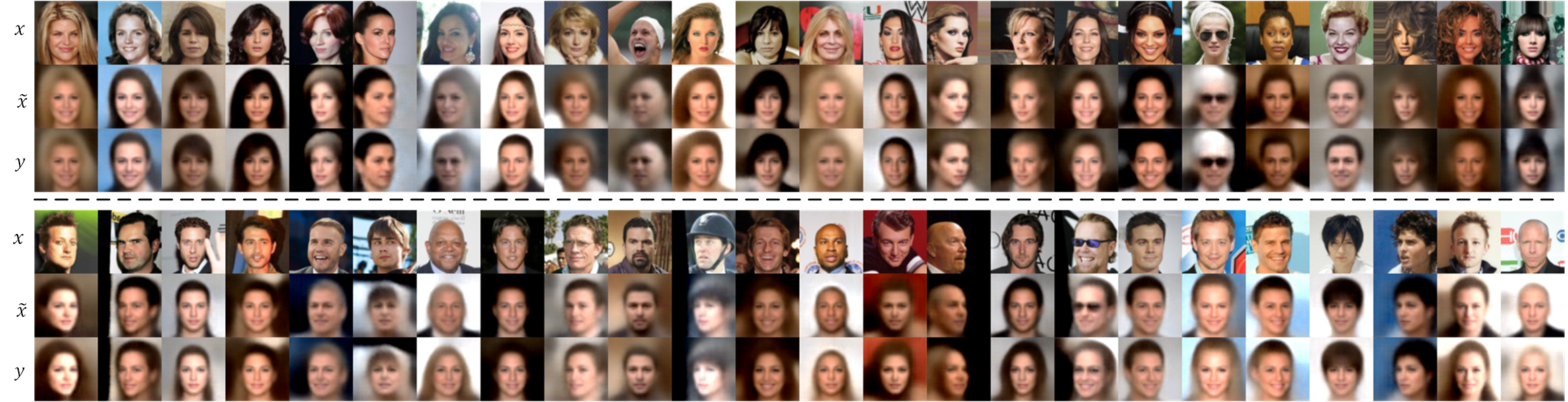}
    \caption{More uncurated results based on $\beta$-VAE. ({\it Top}) Male to Female; ({\it Bottom}) Female to Male. $x$: the input, $\tilde{x}$: the reconstruction, $y$: the translated output.}
    \label{fig:betavae-more}
\end{figure*}

\begin{figure*}[!htbp]
    \centering
    \includegraphics[width=0.99\textwidth]{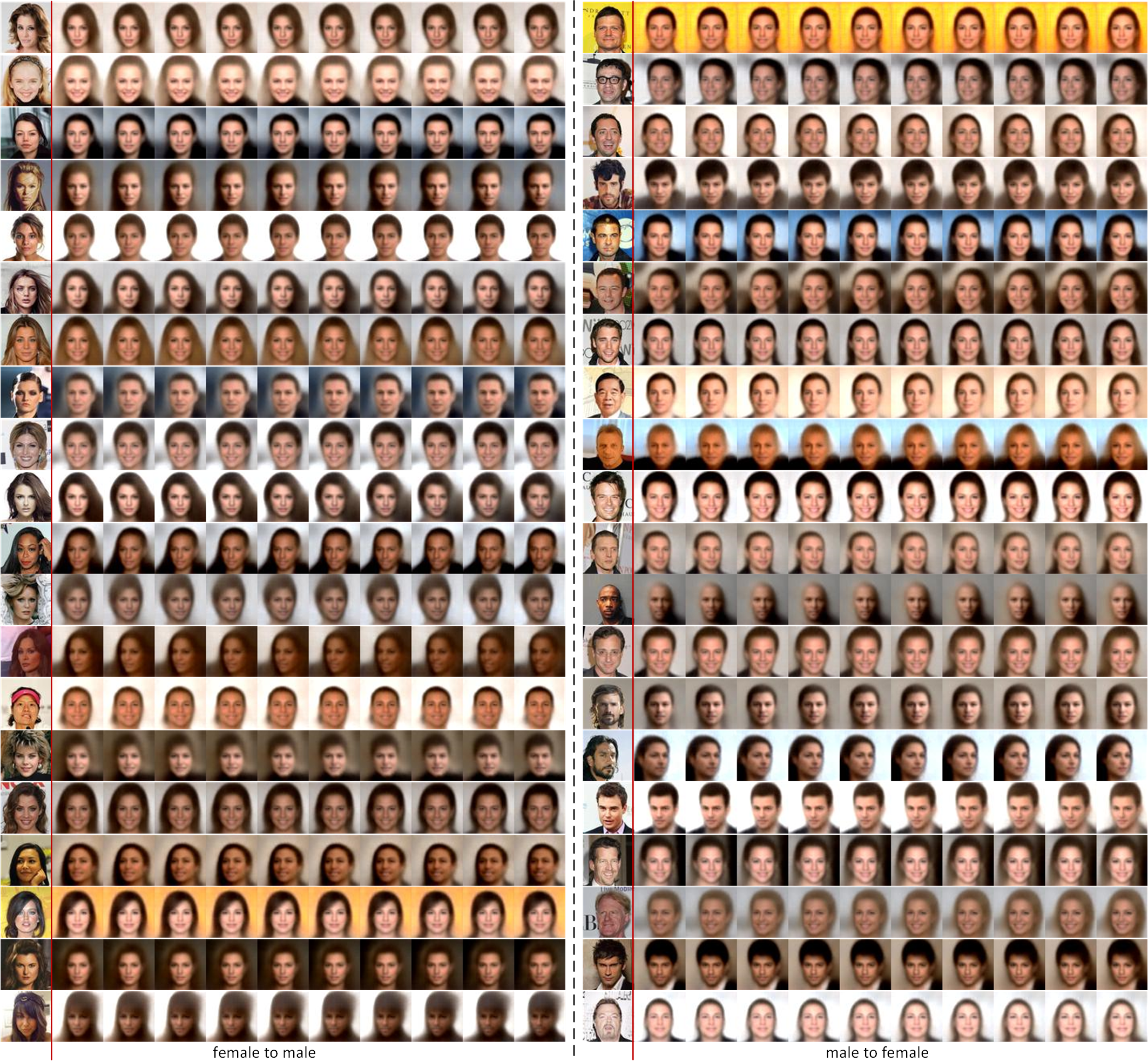}
    \caption{Smooth unpaired image-to-image translation dynamics via MCMC. The leftmost column is the input.}
    \label{fig:betavae-mcmc}
\end{figure*}

We adopt the open-source code in \url{https://github.com/1Konny/Beta-VAE}. We keep all the settings the same but set the latent dimension at 32.
We construct the latent EBM as an one-hidden-layer MLP (32-64-1) and use LeakyReLU for activation. We use SGD for optimization with learning rate 0.1. The MCMC sampler is ran for 10 steps and the step size is 0.1. More results are given in Figure~\ref{fig:betavae-more} and Figure~\ref{fig:betavae-mcmc}.

\section{ALAE}
\begin{figure*}[!htbp]
    \centering
    \includegraphics[width=0.95\textwidth]{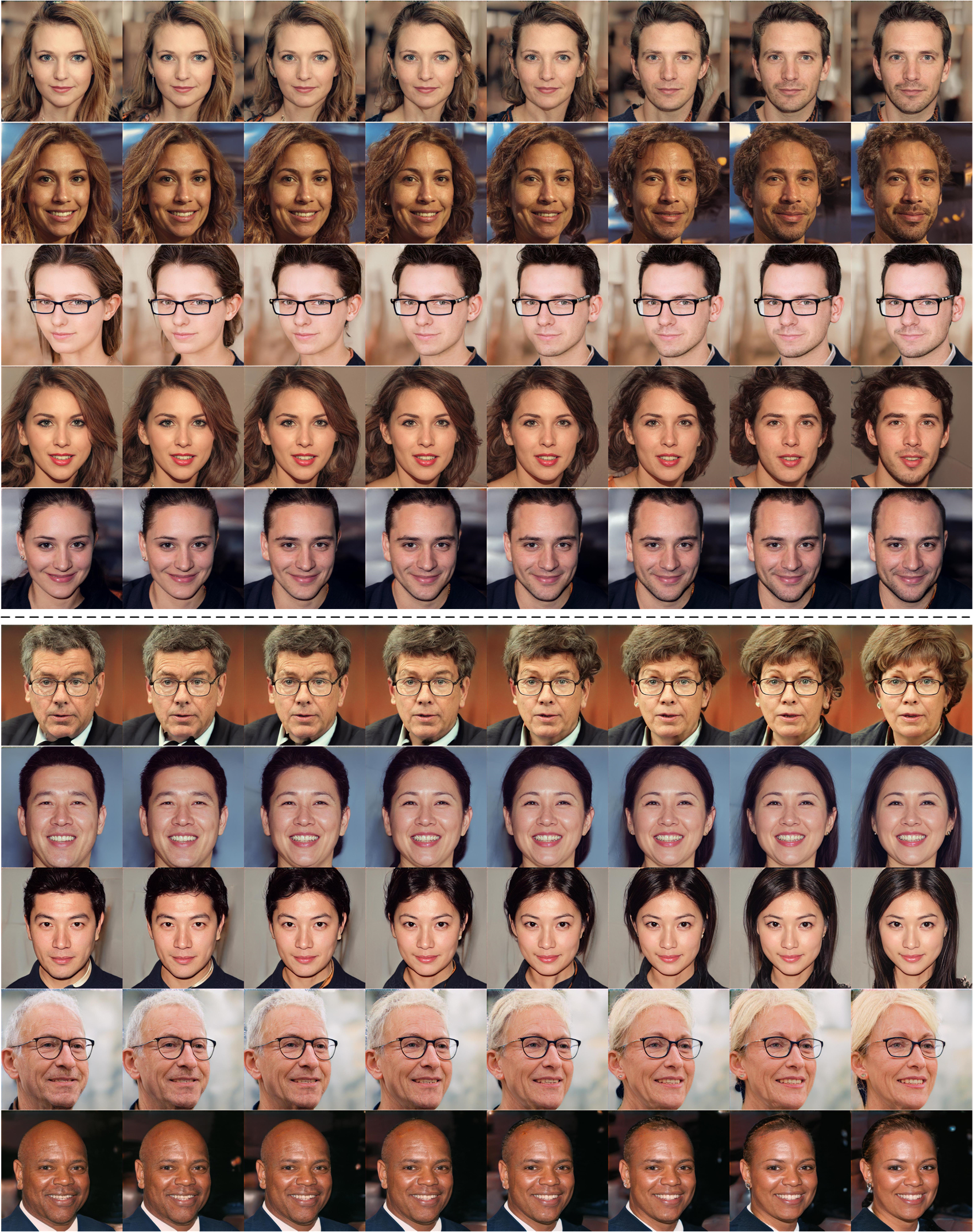}
    \caption{More $1024^2$-pixel image translation dynamics based on ALAE. ({\it Top}) Female to Male, ({\it Bottom}) Male to Female.}
    \label{fig:alae-more-mcmc}
\end{figure*}

We adopt the open-source code in \url{https://github.com/podgorskiy/ALAE} and keep all the settings the same. Implementation details have been given in the main part. More results are given in Figure~\ref{fig:alae-more-mcmc}.

\par \vspace{1mm} \noindent \textbf{Evaluation protocol: } For FID evaluations, we follow the protocol in Appendix C of StarGAN v2, and the public code can be found at {\small \url{github.com/clovaai/stargan-v2}}. Specifically, FID is calculated between translated test images and training images. We report the average FID of each pair of domains. For KID evaluation, we adopt the source code from {\small \url{github.com/taki0112/GAN_Metrics-Tensorflow}}, which has also been used in CF-EBM. 

\section{VQ-VAE-2}
\begin{figure*}[!htbp]
    \centering
    \includegraphics[width=\textwidth]{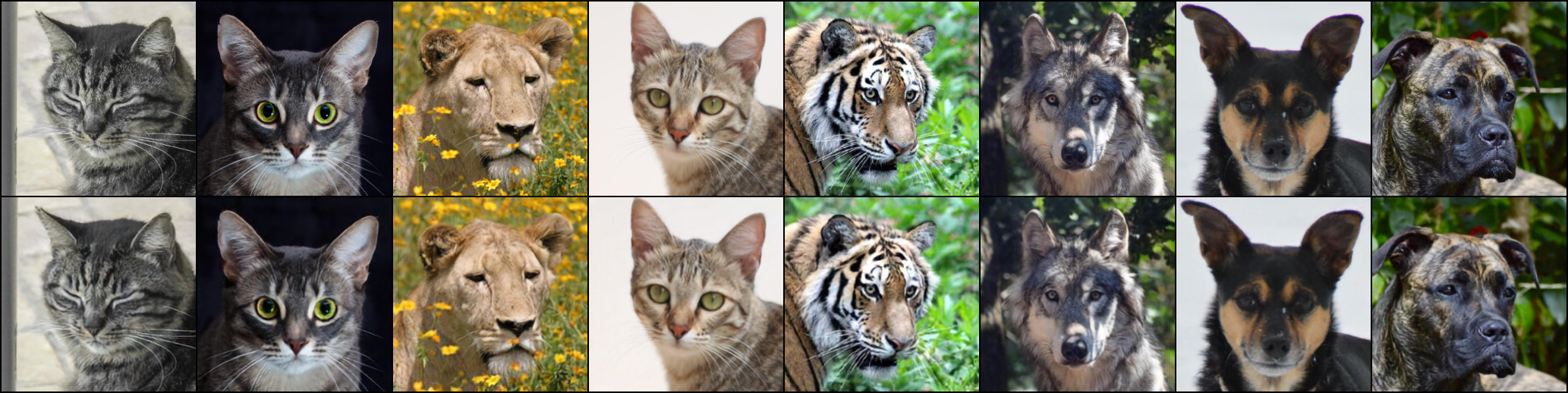}
    \caption{VQ-VAE-2 reconstructions on AFHQ. ({\it Top}) Inputs, ({\it Bottom}) Reconstructions.} 
    \label{fig:vqvae-reconstruction}
\end{figure*}

We adopt the open-source code in \url{https://github.com/rosinality/vq-vae-2-pytorch}. We keep all the settings the same but set the codebook dimesnion at 32 and codebook size at 256. In Table~\ref{tab:mse-vqvae}, we evaluate the reconstruction error when the codebook design varies. Figure~\ref{fig:vqvae-reconstruction} demonstrates the high reconstruction quality on AFHQ. The latent EBM resembles the discriminator of BigGAN~\cite{brock2018large}. We use Adam for optimization where the learning rate is set at 0.001. We run the latent transport for 40 steps with a step size 1.0. We pretrain the VQ-VAE-2 on the whole AFHQ dataset including all the three domains cat, dog and wildlife. Therefore, if we want to obtain a model translating any two domains, the overall efficiency will be even higher than CUT as seen in Table~\ref{tab:cmp-computation}.

\par \vspace{1mm} \noindent \textbf{More comparisons with CF-EBM: } We present more results in Table~\ref{tab:app-cfebm}.
\begin{table}[!htbp]
    \centering
    \scalebox{0.8}{
    \begin{tabular}{c|ccc}
    \toprule
        Datasets & $cat \rightarrow dog$ & $dog \rightarrow cat$ & $vangogh \rightarrow photo$ \\ \midrule
        CF-EBM &  6.20 & 9.21 & \textbf{4.49} \\
        Ours & \textbf{6.01} & \textbf{7.45} & 4.61  \\ \bottomrule
    \end{tabular}
    }
    \vspace{2mm}
    \caption{More KID comparisons with CF-EBM.}
    \label{tab:app-cfebm}
\end{table}

\begin{figure*}[!htbp]
    \centering
    \includegraphics[width=\textwidth]{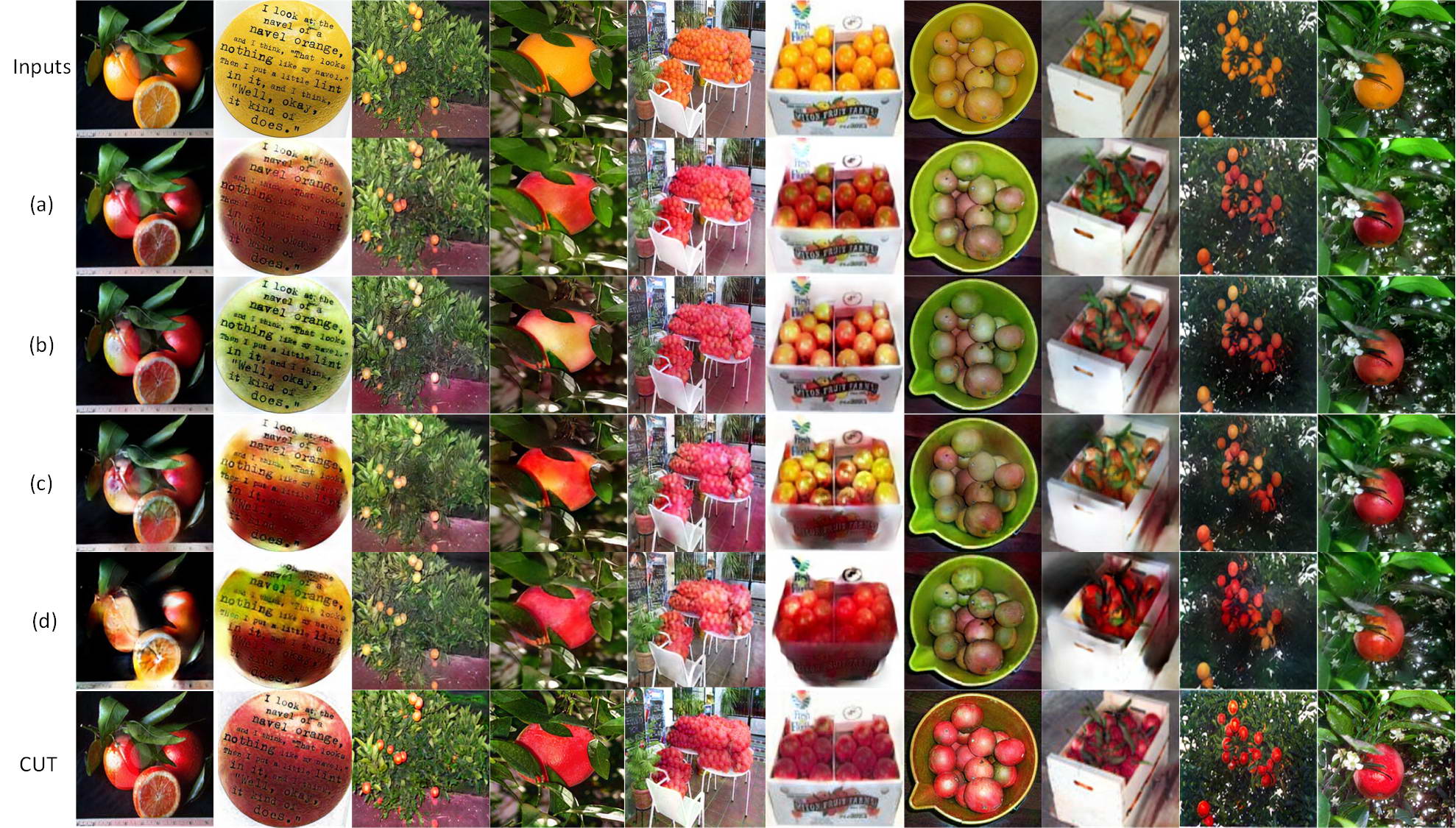}
    \caption{Uncurated translation results on orange $\rightarrow$ apple. (a)-(d) denote different pretraining settings. (a) Pretrain on apple2orange. (b) Pretrain on ImageNet. (c) Pretrain on CelebA-HQ. (d) Pretrain on AFHQ. The last row shows the results from CUT~\cite{park2020contrastive}.}
    \label{fig:vqvae-orange}
\end{figure*}

\begin{figure*}[!htbp]
    \centering
    \includegraphics[width=\textwidth]{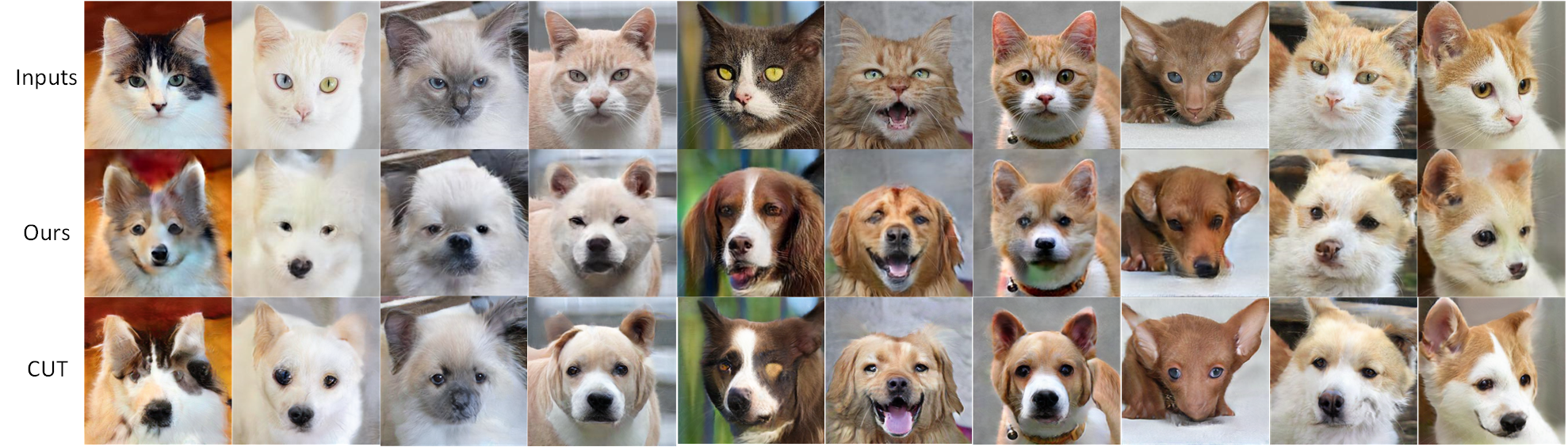}
    \caption{Extended translation results on AFHQ cat $\rightarrow$ dog based on VQ-VAE-2.}
    \label{fig:vqvae-afhq-more}
\end{figure*}

\begin{figure*}[!htbp]
    \centering
    \includegraphics[width=\textwidth]{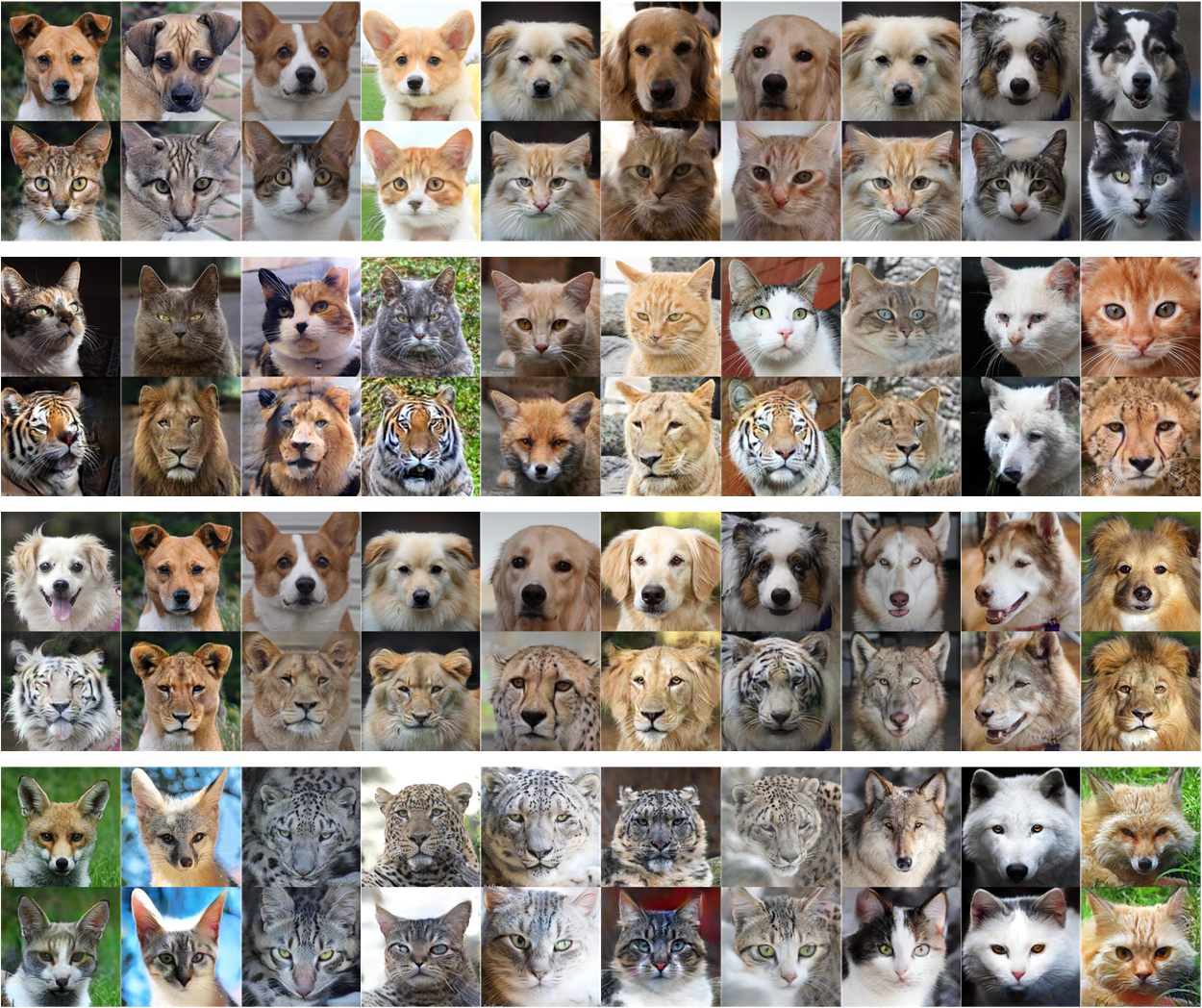}
    \caption{Additional translation results on AFHQ based on VQ-VAE-2. From Top to Bottom: dog $\rightarrow$ cat, cat $\rightarrow$ wild, dog $\rightarrow$ wild, wild $\rightarrow$ cat.}
    \label{fig:vqvae-afhq-dog2cat}
\end{figure*}

\paragraph{More results}
In Figure~\ref{fig:vqvae-orange}, we compare the translation results under various pretraining settings and a baseline model CUT~\cite{park2020contrastive}. We observe although the autoencoder is pretrained with a totally irrelevant dataset, we still can generate reasonable translations. Compared with our standard setting (a) and the baseline CUT, our model demonstrates better style controllability and content preservation ability. Figure~\ref{fig:vqvae-afhq-more} gives extended comparisons on AFHQ cat $\rightarrow$ dog. Figure~\ref{fig:vqvae-afhq-dog2cat} provides additional translation results on AFHQ dog $\rightarrow$ cat, cat $\rightarrow$ wild, dog $\rightarrow$ wild and wild $\rightarrow$ cat.
\begin{table}[!h]
    \centering
    \begin{tabular}{c|c|c|c}
    \toprule
        \diagbox{\textsf{D}}{\textsf{S}} & 128 & 256 & 512 \\ \midrule
        1 & - & 4.73 & 4.59 \\
        2 & - & 2.84 & 2.62 \\
        4 & - & 2.67 & 2.53 \\
        8 & - & 2.40 & 2.38 \\ 
        32 & 2.62 & 2.38 & 2.29 \\ 
        64 & - & 2.31 & 2.07 \\ 
        \bottomrule
    \end{tabular}
    \vspace{2mm}
    \caption{VQ-VAE-2 reconstruction quality (MSE:$10^{-3}$) under various codebook configurations in AFHQ. Each column varies the codebook dimension (\textsf{D}) and each row varies the codebook size (\textsf{S}).}
    \label{tab:mse-vqvae}
\end{table}

\end{document}